\documentclass{article}

\PassOptionsToPackage{numbers, compress}{natbib}

\usepackage[final]{neurips_2021}

\usepackage[utf8]{inputenc} 
\usepackage[T1]{fontenc}    
\usepackage[colorlinks=true,linkcolor=blue,urlcolor=blue,citecolor=purple]{hyperref}       
\usepackage{url}            
\usepackage{booktabs}       
\usepackage{amsfonts}       
\usepackage{nicefrac}       
\usepackage{microtype}      
\usepackage{xcolor}         
\usepackage{physics}
\usepackage[linesnumbered,ruled]{algorithm2e}
\usepackage{graphicx}
\usepackage{caption}
\usepackage{subcaption}

\usepackage{algpseudocode}
\title{Learning Sparsity and Randomness for Data-driven Low Rank Approximation}

\author{%
  Tiejin Chen$^*$\\
  Department of Statistics\\
  University of Michigan\\
  \texttt{tiejin@umich.edu} \\
  \And
  Yicheng Tao$^*$\\
  Department of Computer Science\\
  University of Michigan\\
  \texttt{yctao@umich.edu} \\
}

\begin{document}
\maketitle

\def\thefootnote{*}\footnotetext{Equal contribution.}

\begin{abstract}
Learning-based low rank approximation algorithms can significantly improve the performance of randomized low rank approximation with sketch matrix. With the learned value and fixed non-zero positions for sketch matrices from learning-based algorithms, these matrices can reduce the test error of low rank approximation significantly. However, there is still no good method to learn non-zero positions as well as overcome the out-of-distribution performance loss.

In this work, we introduce two new methods Learning Sparsity and Learning Randomness which try to learn a better sparsity patterns and add randomness to the value of sketch matrix. These two methods can be applied with any learning-based algorithms which use sketch matrix directly. Our experiments show that these two methods can improve the performance of previous learning-based algorithm for both test error and out-of-distribution test error without adding too much complexity.
\end{abstract}

\section{Introduction}
The problem of low-rank approximation (LRA) of matrices is that given a matrix \(A\in \mathbb{R}^{n\times d}\) with \(n\geq d\) and an integer \(0<k\ll d\), find a rank-\(k\) matrix \(A^\prime\) that minimizes the approximation error \(\left\lVert A - A^\prime  \right\rVert _F^2 \) . Truncated Singular value decomposition (SVD) can solve this problem optimally in polynomial time, which is nevertheless too slow for large matrices. To handle this, fast approximate LRA algorithms based on sketching have been proposed ~\cite{koch2007dynamical, cohen2015dimensionality, chierichetti2017algorithms, liberty2007randomized}, which basically construct a sketch matrix \(S\in \mathbb{R}^{m\times n}\) and compute LRA from a much smaller matrix \(SA\) to get a save of both time and storage space. However, these algorithms suffer from a larger approximation error than the optimal LRA, which may affect downstream tasks based on LDA. So the challenge here is to design a sketch matrix that can also produce a good approximation quality.

Data-driven algorithms have been recently developed to handle this challenge \cite{indyk2019learning, ailon2021sparse, liu2020learning, indyk2021few}. They leverage past data that are related to future input as training samples to learn sketching matrices, which result in significant improvement over non-data-driven baselines. However, as mentioned in~\cite{indyk2021few}, the sparsity pattern of the sketching matrix has not been considered to be learned during training, which may additionally improve performance. All the algorithms in the this area consider to learn fixed value for sketch matrix which may cause the problem of over-fitting, and result in poor performance in out-of-distribution situation. Indyk et al.\cite{indyk2019learning} provide a proof of worst case bound which concatenates learned sketch matrix with data-oblivious random matrix vertically. Nonetheless, such kind of method will increase sketch size and get a reduce of efficiency.

In this work, we propose two different methods to overcome the disadvantages we mentioned above. The first method Learning Sparsity will learn the sparsity patterns given target total number of non-zero positions as well as value of sketch matrix $S$. We will use an extra position matrix $D$ to learn the position, and use $S^D = S \odot D$ as final sketch matrix where $\odot$ represents Hardmard product. The second method Learning Randomness tends to learn a Gaussian distribution for every non-zero position of sketch matrix instead of a fixed learned value. We will learn the mean and variance of the distribution. It is easy to notice that our two methods can be easily applied to any learning-based LRA algorithms require sketch matrix $S$ directly. Finally, we combine our two methods to get a new algorithm based on IVY \cite{indyk2019learning}.

Overall our main contribution can be summarized as follows:
 \begin{itemize}
     \item We propose two flexible methods Learning Sparsity and Learning Randomness which can be easily applied to any learning-based LRA method requires sketch matrix $S$ directly.
     \item We combine these two methods to get a new algorithm based on IVY. Our new algorithm can improve the performance of IVY significantly. 
     \item We conduct  experiments to show our method can actually improve the performance of learning-based algorithms.
 \end{itemize}

\section{Related Work}
Low rank approximation with sketch matrix sample from different random distributions \cite{halko2011finding,clarkson2017low,woolfe2008fast,indyk2006stable} has been applied wildly but suffers from high approximation error compared to optimal solution. Learning-based LRA is first introduced in \cite{indyk2019learning} and it proposes IVY, a learning-based LRA algorithm directly follows a common streaming algorithm with random sketch matrix named SCW \cite{clarkson2009numerical,sarlos2006improved}. IVY can increase the performance of sketch methods significantly.  After that, some work try to increase the performance of IVY. Liu et al. \cite{liu2020learning} tends to learn a better sparsity patterns for CountSketch by trying every position after every iteration during training. FewshotSGD \cite{indyk2021few} is proposed to reduce the training time with surrogate loss instead of direct empirical loss used in IVY. Most recently, Liu et al. \cite{liu2022tensor} come up with a tensor-based algorithm with tensor decomposition while Sakaue and Oki \cite{sakaue2022improved} demonstrate iterative hard thresholding method to learn better non-zero positions for sketch matrix.

There are only few work related to the theoretical part of learning-based LRA especially considering error bound. Bartlett et al. \cite{bartlett2022generalization} give a generalization bound for learning-based methods and this result can be improved \cite{sakaue2022improved} by PAC-learning approach \cite{gupta2016pac,gupta2020data}.

\section{Preliminaries}
Given a data matrix $A\in \mathbb{R}^{n\times d}$ and a sketching matrix $S\in \mathbb{R}^{m\times n}$, SCW \cite{clarkson2009numerical} is a widely-used algorithm for sketch-based LDA \cite{indyk2019learning, indyk2021few, liu2020learning}, which is shown in Algorithm \ref{alg:SCW}. It first computes the compact SVD of the sketch $SA$ as $U\Sigma V^\top$, then computes the best rank-$k$ approximation of $AV$ denoted as $[AV]_k$, and finally outputs $[AV]_k V^\top$ as a rank-$k$ approximation of $A$. We remark that SCW is differential if it uses a differentiable SVD algorithm, which is available in PyTorch and used by us.

\begin{algorithm}[H]
    \setcounter{AlgoLine}{0}
    \KwIn{data matrix $A\in \mathbb{R}^{n\times d}$, sketching matrix $S\in \mathbb{R}^{m\times n}$}

    $U, \Sigma, V^\top \gets \text{CompactSVD}(SA) \quad\triangleright\quad \left\{r=\operatorname{rank}(S A), U \in \mathbb{R}^{m \times r}, V \in \mathbb{R}^{d \times r}\right\}$
    
    \KwOut{$[AV]_k V^\top$}
    \caption{SCW}
    \label{alg:SCW}
\end{algorithm}

\section{Method}
We propose two new computational components, namely learning-based sparsity and randomness, for IVY. The learning-based sparsity component provides a solution to the challenge of learning the sparsity pattern of the sketching matrix, while the learning-based randomness component embeds learned Gaussian randomness in the formation of the sketching matrix to facilitate the generalization ability of IVY. We combine each component as well as both of them with IVY and get three new algorithms IVY+LS, IVY+LR, and IVY+LS\&LR.

\subsection{Learning-based Sparsity}
In the previous works of data-driven LDA, the non-zero values of the sketching matrix are learned with their positions fixed, which are randomly chosen before training. This obviously limits the model from learning the best positions for non-zero entries. We fix this issue by learning a sparse matrix $D\in \mathbb{R}_+^{m\times n}$ that indicates the sparsity pattern of the sketching matrix $S$. Specifically, we element-wisely multiply $D$ with $S$ to make $S$ sparse and input the new $S$ to the SCW algorithm. During training, we use an approximation error loss with an $L_1$ regularization on $D$ to control the sparsity of the learned $D$, and if $D$'s entries are less than a threshold $\epsilon$, they will be immediately set to zeros. After $S$ meets our requirement for sparsity, we will stop train $D$ but continue to train $S$. The algorithm of this component combined with IVY, IVY+LS, is shown in Algorithm \ref{alg:IVY+LS}.

\begin{algorithm}[!htbp]
    \setcounter{AlgoLine}{0}
    \KwIn{training set $\{A_1, \cdots, A_N\}\subset \mathbb{R}^{n\times d}$, sketch size $m$, target sparsity $s$, regularization weight $\lambda$, threshold $\epsilon$, learning rate $\eta$}
    
    $S \gets 1^{m\times n}\in \mathbb{R}^{m\times n} \quad\triangleright\quad \{\text{sketching matrix}\}$
    
    $D \gets 1^{m\times n} \in \mathbb{R}_{+}^{m\times n} \quad\triangleright\quad \{\text{sparsity pattern matrix}\}$
    
    \For {$i \gets 1, \cdots, N$}{
        
        $S \gets D\otimes S$

        $L \gets \lVert A_i - \text{SCW}(A_i, S)\rVert_F + \lambda \sum_{i=1}^m\sum_{j=1}^n D_{ij}$

        $S \gets S - \eta\frac{\partial L}{\partial S}$
        
        \If {$\sum_{i=1}^m\sum_{j=1}^n 1_{D_{ij}\neq 0} > ns$}{
            
            $D \gets D - \eta\frac{\partial L}{\partial D}$
            
            $D[D < \epsilon] \gets 0$
        }       
    }
    \KwOut{$S$}
    \caption{IVY+LS}
    \label{alg:IVY+LS}
\end{algorithm}

\subsection{Learning-based Randomness}
IVY directly learns the values of non-zero entries in the sketching matrix $S$, which may lead to the overfitting problem and make $S$ less generalizable. To handle this, we introduce learning-based randomness in the formation of $S$. Specifically, we see every non-zero entry
in $S$ sampled from an independent Gaussian distribution
with learnable mean $\mu$ and variance $\sigma$ and do sampling in every training iteration. By using the reparameterization trick $\mathcal{N}(\mu, \sigma^2) = \sigma\mathcal{N}(0, 1) + \mu$, we can easily update $\mu$ and $\sigma$ during training. Once learned, we can sample every entry of $S$ using $\mu$ and $\sigma$ for inference. The algorithm of this component combined with IVY, IVY+LR, is shown in Algorithm \ref{alg:IVY+LR}. The whole algorithm with both components combined with IVY, IVY+LS\&LR, is shown in Algorithm \ref{alg:IVY+LS&LR}.

\begin{algorithm}[!htbp]
    \small
    \setcounter{AlgoLine}{0}
    \KwIn{training set $\{A_1, \cdots, A_N\}\subset \mathbb{R}^{n\times d}$, sketch size $m$, learning rate $\eta$}
    
    $S \gets 0^{m\times n}\in \mathbb{R}^{m\times n} \quad\triangleright\quad \{\text{sketching matrix}\}$
    
    $\mu, \Sigma \gets 0^{m\times n}\in \mathbb{R}^{m\times n}, 1^{m\times n}\in \mathbb{R}_+^{m\times n}$
    
    \For {$i \gets 1, \cdots, N$}{

        sample $Z \sim \mathcal{N}(0, 1/4)^{m\times n}$
        
        $S \gets Z\otimes \sqrt{\Sigma} + \mu$

        $L \gets \lVert A_i - \text{SCW}(A_i, S)\rVert_F$

        $\mu, \Sigma \gets \mu - \eta\frac{\partial L}{\partial \mu}, \text{relu}(\Sigma - \eta\frac{\partial L}{\partial \Sigma})$
        }
    \KwOut{$S$}
    \caption{IVY+LR}
    \label{alg:IVY+LR}
\end{algorithm}

\begin{algorithm}[!htbp]
    \small
    \setcounter{AlgoLine}{0}
    \KwIn{training set $\{A_1, \cdots, A_N\}\subset \mathbb{R}^{n\times d}$, sketch size $m$, target sparsity $s$, regularization weight $\lambda$, threshold $\epsilon$, learning rate $\eta$, early stop flag $flag$}
    
    $S \gets 0^{m\times n}\in \mathbb{R}^{m\times n} \quad\triangleright\quad \{\text{sketching matrix}\}$
    
    $D, \mu, \Sigma \gets 1^{m\times n} \in \mathbb{R}_{+}^{m\times n}, 0^{m\times n}\in \mathbb{R}^{m\times n}, 1^{m\times n}\in \mathbb{R}_+^{m\times n} \quad\triangleright\quad \{D: \text{sparsity pattern matrix}\}$
    
    \For {$i \gets 1, \cdots, N$}{

        sample $Z \sim \mathcal{N}(0, 1/4)^{m\times n}$
        
        $S \gets D\otimes (Z\otimes \sqrt{\Sigma} + \mu)$

        $is\_sparse \gets \sum_{i=1}^m\sum_{j=1}^n 1_{D_{ij}\neq 0} > ns$

        \If {is\_sparse} {
            $\lambda \gets 0$
        }

        $L \gets \lVert A_i - \text{SCW}(A_i, S)\rVert_F + \lambda \sum_{i=1}^m\sum_{j=1}^n \lvert D_{ij}\rvert$

        $\mu, \Sigma \gets \mu - \eta\frac{\partial L}{\partial \mu}, \text{relu}(\Sigma - \eta\frac{\partial L}{\partial \Sigma})$

        \If {$is\_sparse$ \& $flag$}{
            
            $D \gets D - \eta\frac{\partial L}{\partial D}$
            
            $D[D < \epsilon] \gets 0$
        }
    }
    \KwOut{$S$}
    \caption{IVY+LS\&LR}
    \label{alg:IVY+LS&LR}
\end{algorithm}

\section{Experimental results}
\subsection{Setting}
It can be noticed that our two methods Learning Sparsity(LS) and Learning Randomness(LR) can be applied to any learning-based algorithms requires sketch matrix $S$ directly besides IVY. In other word, our method can be encode to many algorithms except Butterfly\cite{ailon2021sparse} which does not use $S$ directly. However, as mentioned in the previous section, we will only encode our methods to IVY to prove the validity.

\textbf{Dataset} We will only use \textbf{Hyper} dataset from ~\cite{imamoglu2018hyperspectral} as our training set. \textbf{Hyper} contains hyperspectral images from natural scenes and each matrix has size of $1024 \times 768$. We will randomly sample 500 images as training set while use 100 images as test set. We will also use \textbf{Fish} dataset$^{\dagger}$ from Kaggle as an unseen test set during whole training process to verify generalization ability and out-of-distribution performance of our methods. \textbf{Fish} contains the images from different classes of fish. We randomly choose 100 images with size of  $1024 \times 768$ and transform them into grayscale images.

\textbf{Sparsity and Density} Overall, one measure to describe the sparsity of one matrix is total number of non-zero entries. However, here we will define density to describe the sparsity of sketch matrix. More specifically, density means total number of non-zero positions in one column for sketch matrix. CountSketch is a matrix with density1 and a matrix with density equal to its sketch size is a dense matrix. 

\textbf{Metric} For the test error, we follow the instruction from IVY \cite{indyk2019learning} and fewshotSGD\cite{indyk2021few} and use a test error as:
\begin{equation}
    err_{A}(S) = \lVert A - SCW(A,S) \rVert_F^2 -  \lVert A - A_k \rVert_F^2
\end{equation}
Where $A_k$ is the optimal rank k matrix respect to $A$ and it can be obtained by truncated SVD. 

In all experiments, we will use target rank is 10 which is from the original paper from IVY. As for training strategies, we follow the original strategy from IVY with learning rate equals to 1 and momentum equals to 1. The threshold $\epsilon$ is set to 0.5 and parameter $\lambda$ for $L_1$ regularization is set to $0.0003$ when density is 1 and $0.0001$ when density is larger than 1. All the algorithms are implemented by PyTorch 1.13.0.

\def\thefootnote{$\dagger$}\footnotetext{can be found at https://www.kaggle.com/datasets/crowww/a-large-scale-fish-dataset?resource=download}

\subsection{Average Test Error}
Firstly, we test all our method as well as IVY on the \textbf{Hyper} with different $m$. For Learning Sparsity, we can not direct control density of sketch matrix while we can control the sparsity e.g the total number of non-zero entries in sketch matrix. We will report the test error for sketch matrix learned by Learning Sparsity with less sparsity than other methods. We run the each setting for 5 times and report the average test error. The results are shown in Table \ref{tab:1}.

As we can see, Learning Sparsity can increase the performance of IVY significantly regardless of $m$. It seems that our method can learn a better sparse pattern than random choice. On the other hand, Learning Randomness cannot perform very well when sketch size is small but do improve the performance when sketch size is more reasonable. After combining Learning Sparsity and Learning Randomness, the performance will further increase during most situation. However, with very small sketch size, the disadvantage of randomness is still dominating.

\begin{table}[!htbp]
\centering
\caption{Test error with density1}
\label{tab:1}
\begin{tabular}{|c|c|c|c|c|} 
\hline
$k$, $m$   & IVY  & IVY+LS & IVY+LR & IVY+LS\&LR \\ \hline
10, 10 & 4.12 & 3.59   & 6.20   & 5.07      \\ \hline
10, 20 & 1.14 & 0.95   & 1.55   & 0.86      \\ \hline
10, 40 & 0.43 & 0.23   & 0.41   & 0.21      \\ \hline
10, 80 & 0.12 & 0.07   & 0.11   & 0.06       \\ \hline
\end{tabular}
\end{table}

\begin{figure}
     \centering
     \begin{subfigure}[b]{0.49\textwidth}
         \centering
         \includegraphics[width=\textwidth]{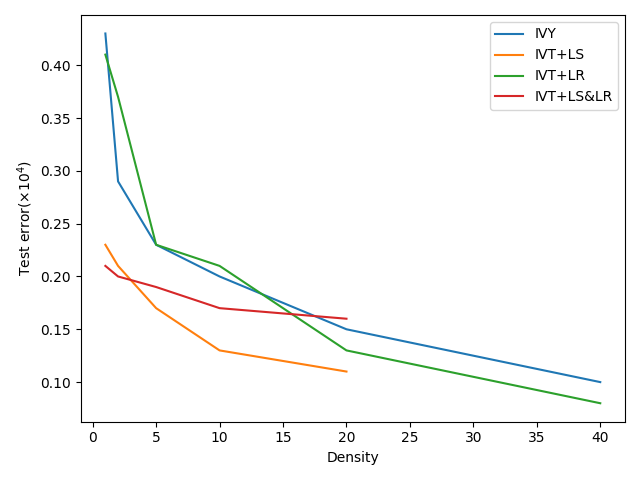}
         \caption{Density 1}
         \label{fig:D}
     \end{subfigure}
     \hfill
     \begin{subfigure}[b]{0.49\textwidth}
         \centering
         \includegraphics[width=\textwidth]{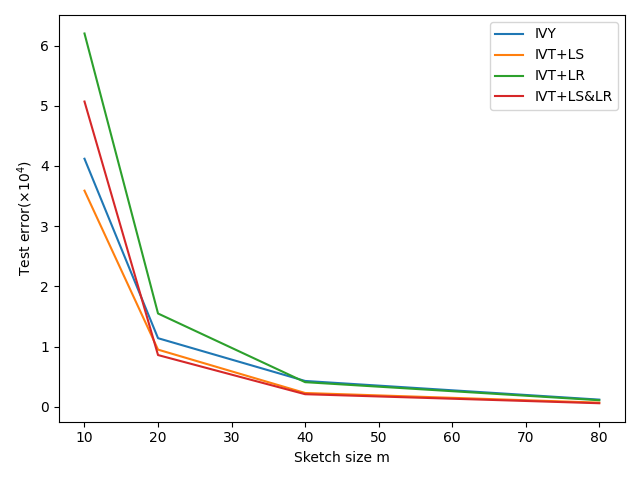}
         \caption{Density 20}
         \label{fig:M}
     \end{subfigure}
     \hfill
\caption{Test error according to training time}

\end{figure}

Now let us take a look at the influence of different density. In this experiment, we will fix $k$ and $m$ to be 10 and 40 respectively. Again, for Learning Sparsity and combined method, once the sketch matrix meets the requirement of sparsity we will stop train $D$. We will run in total 500 iterations for the experiments. Our result are shown in Table \ref{tab:2}. Overall, Learning Sparsity can actually learn a better sparse patterns while Learning Randomness can also work well in some situations. We visualize the results in Figure \ref{fig:D} and Figure \ref{fig:M}.

\begin{table}[!htbp]
\centering
\caption{Test error with different density based on sketch size 40}
\label{tab:2}
\begin{tabular}{|c|c|c|c|c|} 
\hline
Density   & IVY  & IVY+LS & IVY+LR & IVY+LS\&LR \\ \hline
1& 0.43 & 0.23   & 0.41   & 0.21      \\ \hline
2 & 0.29  & 0.21   &  0.37  &  0.20     \\ \hline
5 & 0.23  & 0.17   &  0.23  &  0.19    \\ \hline
10 & 0.20 & 0.13   &  0.21  &  0.17      \\ \hline
20 & 0.15 & 0.11   &  0.13  &  0.16      \\ \hline
40 & 0.10 & -   & 0.08  & -       \\ \hline
\end{tabular}
\end{table}

\subsection{Running Time}
\begin{figure}
     \centering
     \begin{subfigure}[b]{0.49\textwidth}
         \centering
         \includegraphics[width=\textwidth]{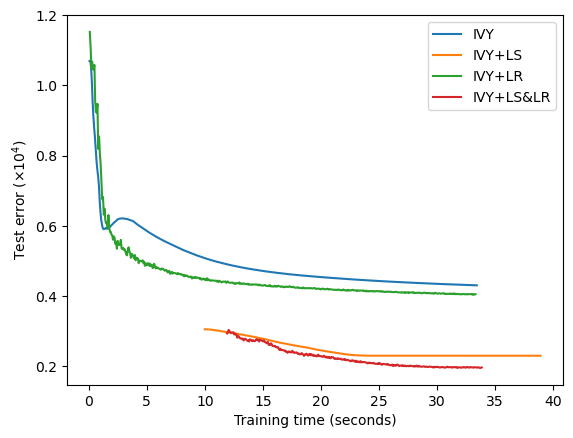}
         \caption{Density 1}
         \label{fig:D1}
     \end{subfigure}
     \hfill
     \begin{subfigure}[b]{0.49\textwidth}
         \centering
         \includegraphics[width=\textwidth]{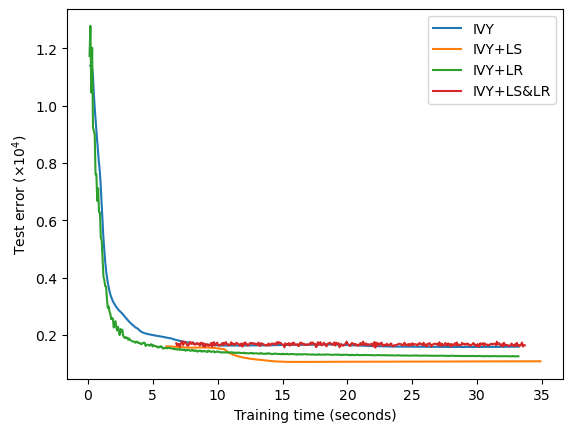}
         \caption{Density 20}
         \label{fig:D20}
     \end{subfigure}
     \hfill
\caption{Test error according to training time}

\end{figure}

The inference time of our methods will be same as IVY or random matrix since the output of our methods will be no different compared to them. However, training time is also considered as one metric to measure the algorithms in the area. Here we report the training time of 500 iterations 
for all the methods including IVY with $m=40$, density$=1$ in Table \ref{tab:3}. We can find that our two new methods will not affect too much about training time. We also visualize our result for $m=40$, density$=1$ and $m=40$, density $=20$ in Figure \ref{fig:D1} and Figure \ref{fig:D20}. Noticing that before sketch matrix from Learning Sparsity meet our sparsity requirement, the comparison between Learning Sparsity and other method is unfair. That is the reason why the data for Learning Sparsity (and combined method) is not start from zero. From the figures we can easily see that our methods are better even considering training time instead of iteration numbers.
\begin{table}[!htbp]
\centering
\caption{Training time of 500 iterations}
\label{tab:3}
\begin{tabular}{|c|c|c|c|c|} 
\hline
IVY  & IVY+LS & IVY+LR & IVY+LS\&LR \\ \hline
26.78s & 26.99s  & 26.84s   & 27.01s      \\ \hline
\end{tabular}
\end{table}

\subsection{Out-of-distribution Test Error}
\begin{figure}
     \centering
     \begin{subfigure}[b]{0.49\textwidth}
         \centering
         \includegraphics[width=\textwidth]{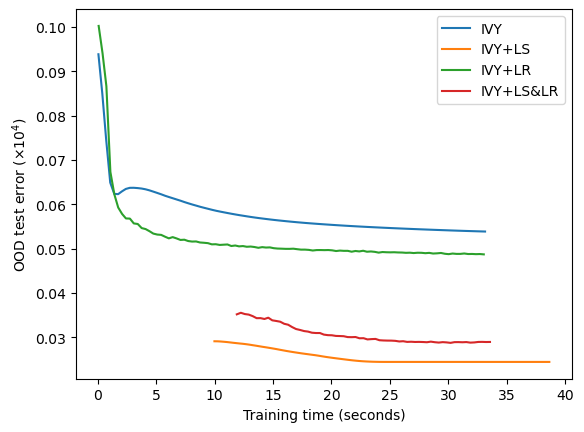}
         \caption{Density 1}
         \label{fig:D1_fish}
     \end{subfigure}
     \hfill
     \begin{subfigure}[b]{0.49\textwidth}
         \centering
         \includegraphics[width=\textwidth]{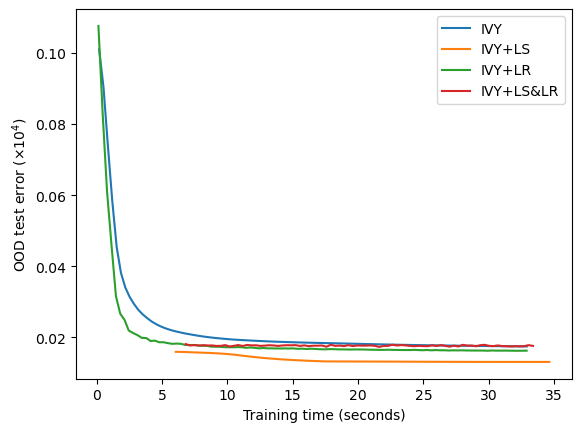}
         \caption{Density 20}
         \label{fig:D20_fish}
     \end{subfigure}
     \hfill
\caption{OOD Test error according to training time}

\end{figure}

In this section, we will report our results for the unseen test set \textbf{Fish} to verify the generalization ability for our methods.  We will apply the best sketch matrix which means it can reach the minimum test error on \textbf{Hyper} to \textbf{Fish} and get the test error for \textbf{Fish} on $m=40$. The results are shown in Table \ref{tab:4}. We also keep track of test error for $\textbf{Hyper}$ and the results are shown in Figure\ref{fig:D1_fish} and Figure \ref{fig:D20_fish}. The results proves that our methods can improve out-of-distribution performance. However we need to be careful if we use combined algorithm. And we can see from figures, out-of-distribution test error will decrease with increasing of training time which prove that learning-based algorithms have generalization ability.

\begin{table}[!htbp]
\centering
\caption{\textbf{Fish} Test error ($\times 10^3$) with different density based on sketch size 40}
\label{tab:4}
\begin{tabular}{|c|c|c|c|c|} 
\hline
Density   & IVY  & IVY+LS & IVY+LR & IVY+LS\&LR \\ \hline
1& 0.52 & 0.24   & 0.47   & 0.28      \\ \hline
2 & 0.37  & 0.24   & 0.45   & 0.26      \\ \hline
5 & 0.29  & 0.19   & 0.31   & 0.24     \\ \hline
10 & 0.24 & 0.16   & 0.26   & 0.21       \\ \hline
20 & 0.18 & 0.13   & 0.16   & 0.18       \\ \hline
40 & 0.09 & -   & 0.08  & -       \\ \hline
\end{tabular}
\end{table}

\section{Conclusion and Further Work}
In this paper we propose two new methods to learn the sparsity and randomness for sketch matrix separately and provide a new algorithm combining this two methods. It is not hard to notice that both two method can be encode to any learning-based LRA algorithm with explicit use of sketch matrix. Our experiments shows our methods can improve the performance of original IVY. However, in our paper we do not give any theoretical result for neither of our new method, which we leave for further work. Also, combining these two methods does not necessarily lead to better performance than each of them, especially when the sketch size is large, which is another interesting direction for research.

\newpage
\bibliographystyle{plainnat}
\bibliography{ref}

\end{document}